\documentclass[pdflatex,sn-mathphys-num]{sn-jnl}

\usepackage{graphicx}%
\usepackage{multirow}%
\usepackage{amsmath,amssymb,amsfonts}%
\usepackage{amsthm}%
\usepackage{mathrsfs}%
\usepackage[title]{appendix}%
\usepackage{xcolor}%
\usepackage{textcomp}%
\usepackage{manyfoot}%
\usepackage{booktabs}%
\usepackage{algorithm}%
\usepackage{algorithmicx}%
\usepackage{algpseudocode}%
\usepackage{listings}%

\theoremstyle{thmstyleone}%
\theoremstyle{thmstyletwo}%

\theoremstyle{thmstylethree}%

\begin{document}

\title[Article Title]{ABCO: Adaptive Bacterial Colony Optimisation}


\author{\fnm{Barisi} \sur{Kogam}}\email{barisikogam@outlook.com}

\author*[1]{\fnm{Yevgeniya} \sur{Kovalchuk}}\email{y.kovalchuk@ucl.ac.uk}

\author[2]{\fnm{Mohamed Medhat} \sur{Gaber}}\email{mohamed.gaber@bcu.ac.uk}

\affil[1]{\orgdiv{UCL Centre for Advanced Research Computing}, \orgname{University College London}, \orgaddress{\country{UK}}}

\affil[2]{\orgdiv{School of Computing and Digital Technology}, \orgname{Birmingham City University}, \orgaddress{\country{UK}}}


\abstract{This paper introduces a new optimisation algorithm, called Adaptive Bacterial Colony Optimisation (ABCO), modelled after the foraging behaviour of E. coli bacteria. The algorithm follows three stages--explore, exploit and reproduce---and is adaptable to meet the requirements of its applications. The performance of the proposed ABCO algorithm is compared to that of established optimisation algorithms--particle swarm optimisation (PSO) and ant colony optimisation (ACO)--on a set of benchmark functions. Experimental results demonstrate the benefits of the adaptive nature of the proposed algorithm: ABCO runs much faster than PSO and ACO while producing competitive results and outperforms PSO and ACO in a scenario where the running time is not crucial.}




\maketitle

\section{Introduction}\label{sec1}

Optimisation--the process of finding an optimal solution for a given input--underpins many scientific and engineering solutions such as network scheduling and image processing \cite{GHAREHCHOPOGH2019}. Many well-established optimisation algorithms are designed to mimic behaviours found in nature. For example, the particle swarm optimisation (PSO) algorithm \cite{kennedy1995} is modelled after the flocking behaviour of birds and fish, while the ant colony optimisation (ACO) algorithm \cite{dorigo1996} depicts the foraging behaviour of the forager ants in an ant colony. One of the limitations of existing optimisation algorithms mimicking swarm behaviour is their computational cost--they rely on many iterations of computation steps performed over a large set of particles (population of individuals). To address this limitation, this study investigates the potential of leveraging the principle of the explore–-exploit trade-off when designing a swarm optimisation algorithm. In particular, we propose a novel optimisation algorithm that is adaptable to the task at hand by balancing the trade-off between speed and accuracy. The proposed algorithm, called Adaptive Bacterial Colony Optimisation (ABCO), takes the foraging behaviour of E. coli bacteria as the basis, similar to the previously proposed BCO algorithm \cite{niu2012}, and augments it by introducing two modes of bacteria movement: exploration and exploitation. Testing ABCO on a set of benchmark functions demonstrates its ability to find optimal solutions much faster than traditional swarm optimisation algorithms such as PSO and ACO by enabling and balancing the exploration of the search space and exploiting promising search paths.

The rest of this paper is organised as follows. Section 2 discusses related work. Section 3 provides the formal description of the proposed ABCO algorithm. Section 4 outlines the experimental setup for comparing ABCO with PSO and ACO, while Section 5 details the experimental results. Section 6 concludes the paper.

\section{Related work}\label{sec2}

Optimisation algorithms all have one goal: to find the optimum solution to a given problem (in practice, finding the maximum or minimum value of a function formalising the problem). The application of these algorithms in the real world is typically seen in the Engineering and Artificial Intelligence fields, as these algorithms solve complex engineering design optimisation problems \cite{yang2010} with great accuracy. Many optimisation algorithms are designed after behaviours observed in nature. For example, PSO \cite{kennedy1995} takes its inspiration from the flocking of birds together to communicate with each other; the Bees algorithm \cite{pham2006} -- from the foraging behaviours of bees in a bee colony; the firefly algorithm \cite{yang2009} -- from fireflies searching for the location with the best brightness; the bacterial foraging optimization (BFO) \cite{passino2002} and bacterial colony optimisation (BCO) algorithms \cite{niu2012} -- from the foraging behaviour of E. coli bacteria. 

Despite the number and diversity of the proposed nature-inspired optimisation algorithms, all come with their own limitations. For example, the authors of PSO \cite{kennedy1995} highlight an overshooting problem that causes the boids (birds) to explore outside the search space. The lack of a communication mechanism between bacteria in BFO \cite{passino2002} limits its ability to find an optimal solution quickly and accurately. The high number of tunable parameters in the Bees algorithm \cite{pham2006} makes it unattractive for real-world applications. To overcome these limitations, this study proposes a novel optimisation algorithm that balances exploration and exploitation behaviours to enhance the convergence speed, a constraint mechanism to prevent overshooting, and an adequate number of tunable parameters. 

More specifically, this study takes an inspiration from and improves upon the recent BCO algorithm \cite{niu2012}. The authors of the original BCO algorithm modelled it around the life-cycle of the E-coli bacteria, which includes such stages as chemo-taxis, communication, elimination, reproduction and migration. In the chemo-taxis stage, bacteria move randomly in search for the best solution. The authors refer to these movements as running and tumbling (changing direction). Following chemo-taxis, the bacteria communicate with each other to discover which bacterium has found the better solution. The authors indicate that the chemo-taxis and communication stages are run together as the bacteria need communication to help direct their tumble movement. The authors created three models of communication to improve the chemo-taxis of individual Bacteria. The first model is dynamic neighbour-oriented communication, which involves the communication of bacteria with their respective neighbours. The second model is random-oriented communication, which involves the communication of bacteria randomly. The third model is group-oriented communication, which involves the communication of bacteria in groups. Following chemo-taxis and communication, the next stage of the BCO algorithm is elimination, which involves the removal of the bacteria that have found a poor solution. The bacteria survived after the elimination stage are reproduced to create new bacteria in the reproduction stage. In the migration stage, bacteria are allowed to extend their current search space in attempt to find better solutions. 

The proposed ABCO algorithm optimises the original BCO algorithm \cite{niu2012} by reducing the number of stages while balancing exploration and exploitation strategies when looking for the best solution. In particular, instead of relying on computationally expensive communication mechanisms and risky migration, bacteria in the proposed ABCO algorithm simply account for the solutions of their neighbours to optimise their own movement. Furthermore, ABCO takes a slightly different approach to reproduction: the algorithm generates new bacteria by making use of the weighted-sum average of the neighbours of top-performing bacteria.

Overall, for the first time, this study investigates the potential of modelling exploration and exploitation behaviours as a way of balancing the trade-off between speed and accuracy when finding optimal solutions. The adaptive nature of the proposed ABCO algorithm achieved through tuning the ratio between exploration and exploitation depending on the application requirements differentiates it from the existing swarm optimisation algorithms.

\section{Proposed ABCO algorithm}\label{method}
\begin{figure}
\centering
\includegraphics[width=8cm, height=15cm]{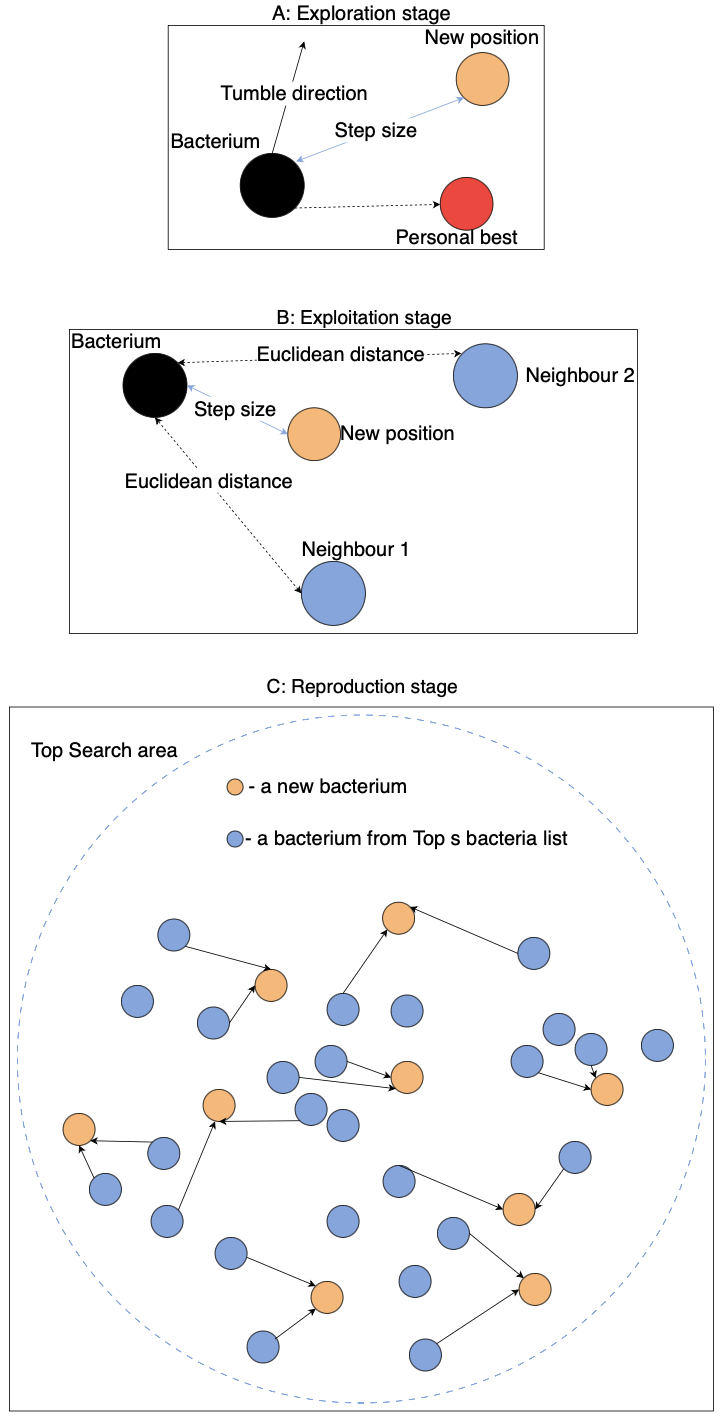}\\
\caption{The proposed ABCO algorithm illustration: (a) exploration stage; (b) exploitation stage; (c) reproduction stage.}\label{fig:illustration}
\end{figure}

Figure \ref{fig:illustration} illustrates the three stages of the algorithm (exploration, exploitation and reproduction), while their pseudocodes are listed in Algorithms \ref{algo:algo_explore}, \ref{algo:algo_exploit} and \ref{algo:algo_repro}, respectively. The tuneable parameters of the ABCO algorithm are detailed in Table \ref{table:ABCOParamDesc}, and its implementation can be found on GitHub: \href{https://github.com/Brzy02/ABCO-Algorithm}{https://github.com/Brzy02/ABCO-Algorithm}.

First, bacteria are dispersed randomly across the search space (Algorithm \ref{algo:algo_explore}, line 4). This is done according to the test function: its upper and lower bounds, as well as the number of dimensions. For example, if the test function is one-dimensional, then the bacteria will only have an x position in the search space; if the test function is two-dimensional, then the bacteria will have x and y positions; and so on. Once the bacteria are seeded across the search space, the algorithm starts its run-time.

The first, exploration stage uses a tuneable parameter $N_{explor}$ (see Table \ref{table:ABCOParamDesc}), which directs how many times the stage is run. Within the exploration stage, there is a tumble step, where the bacteria tumble stochastically across the search space. The tumble step is controlled via a tuneable parameter $N_{tum}$, which sets the number of times the bacteria tumble per the exploration stage. To prevent over-shooting (bacteria going outside the search space), a mechanism is put in place that checks each bacterium’s position every tumble step and corrects it if needed to ensure search is performed within the test function's constraints (Algorithm \ref{algo:algo_explore}, lines 11-16). The next step checks if the bacterium has found a better solution in the search space than its personal best found in the previous runs, which is defined by whether the problem is a maximisation or minimisation problem (Algorithm \ref{algo:algo_explore}, lines 17-36). If the bacterium’s new solution is better than its personal best, the new solution becomes the bacterium's personal best. The bacterium is allowed to move towards the new personal best position if the difference between the best and current positions exceeds a tuneable threshold $e$ to prevent unnecessary micro-movements for small gains (Algorithm \ref{algo:algo_explore}, lines 21 and 31). If the difference is below the threshold, then the bacterium continues its random movement across the search space (Algorithm \ref{algo:algo_explore}, lines 23 and 33). At the end of the exploration stage, all the bacteria, with their respective solution and position, carry on to the next, exploitation stage.

\begin{table}
\tiny
\caption{Tuneable parameters of the proposed ABCO algorithm}
\centering
\begin{tabular}{||r l||} 
 \hline
 Parameter & Description \\ [0.5ex] 
 \hline\hline
$size$ & population size\\
$p$ & population \\
$iter$ & iteration\\
$N_s$ & step size\\
$N_{explor}$ & number of explore steps\\
$N_{explt}$ & number of exploit steps\\
$N_{tum}$ & number of tumble steps\\
$tumbledirection$ & signifies the stochastic direction a bacterium tumbles\\
$distance$ & the Euclidean distance between two bacteria positions\\
$dim$ & dimension of function being optimized\\
$lb$ & lower bound of search space\\
$ub$ & upper bound of search space\\
$e$ & fitness threshold\\
$s$ & population split\\
$k$ & number of neighbours\\
$b\in p$ & bacterium in population \\
$b.position$ & position of bacterium in search space\\
$b.solution$ & solution found by bacterium in search space\\
$b.bestposition$ & best position of bacterium in search space\\
$b.bestsolution$ & best solution found by bacterium \\   
$mode\{"min", "max"\}$ & defines if the optimisation is minimization or maximization\\
$function$ & fitness function\\
$b.previousbestsolution$ & best solution found by bacterium in previous iteration \\   
$generationgap$ & \% of $iter$ to wait before checking for changes in bacteria solutions \\
$unchangedThreshold$ & \% of bacteria whose solutions didn't change since previous check
 \\ [1ex] 
 \hline
\end{tabular}
\label{table:ABCOParamDesc}
\end{table}

\begin{algorithm} 
\begin{algorithmic}[1] 
\caption{ABCO algorithm: initiation and exploration step}\label{algo:algo_explore}
\tiny
\Require{$function$, $dim$, $lb$, $ub$, $p$, $size$, $iter$, $b\in p$, $b.position$, $b.solution$, $b.bestsolution$, $b.bestposition$ , $N_s$, $N_{explor}$, $N_{explt}$, $N_{tum}$, $tumbledirection$, $distance$, $e$, $s$, $k$, $mode$.}

\Ensure{$Global Best$ (Best solution found amongst bacteria)}
\Statex
\Function{ABCO}{$\mathbf{function, p, iter, N_s, N_{exp}, N_{explt}, N_{tum}, dim, lb, ub, e, s, k,
mode}$}
    \State {$GlobalBest$ $\gets$ {$0$}}
    \State {$Initialise$ $parameters$}
    \State {$Seed$ $bacteria$ $in$ $the$ $search$ $space$ $randomly$ $according$ $to$ $function$, $dim$, $lb$, $ub$}
    \For{$i \gets 1$ to $iter$}  
    \For{$ explor \gets 1$ to $N_{explor}$}\Comment{$Exploration$ $step$}
    \For{$ j \gets 1$ to $N_{tum}$}
    \For{$ b $ $ in $ $ p$}  \Comment For each bacterium in population
    \State {$b.position$ $\gets$ {$b.position + N_s*tumbledirection$}} \Comment move bacteria in search space
    \\
    
    \If{$b.position$ $<$ $lb$}
    \State{$b.position \gets move$ $b.position$ $randomly$ $in$ $the$ $search$ $space$ $according$ $to$ $function$, $dim$, $lb$, $ub$} \Comment prevent bacterium from over-shooting
    \EndIf
    \If{$b.position$ $>$ $ub$}
    \State{$b.position \gets move$ $b.position$ $randomly$ $in$ $the$ $search$ $space$ $according$ $to$ $function$, $dim$, $lb$, $ub$}
    \EndIf
    \If{$mode \gets "min"$}
        \If {$b.solution$ $<$ $b.bestsolution$}
    \State{$b.bestsolution \gets b.solution$}
    
     \If {$b.solution-b.bestsolution$ $>$ $e$}
    \State{$b.position \gets b.position+(N_s*distance(b.position, b.bestposition))$} \Comment Move to the location of personal best
    \Else 
    \State{$b.position \gets b.position+(N_s*tumbledirection)$} \Comment Move randomly in search space
    
    \EndIf
    \EndIf
    \EndIf
    \If{$mode \gets "max"$}
        \If {$b.solution$ $>$ $b.bestsolution$}
    \State{$b.bestsolution \gets b.solution$}
    
     \If {$b.solution-b.bestsolution$ $>$ $e$}
    \State{$b.position \gets b.position+(N_s*distance(b.position, b.bestposition))$} \Comment Move to the location of personal best
    \Else 
    \State{$b.position \gets b.position+(N_s*tumble direction)$} \Comment Move randomly in search space
    
    \EndIf
    \EndIf
    \EndIf
    \EndFor
    \EndFor
    \EndFor
    \Comment{$Exploitation$ $step$}\\
    \Comment{$Reproduction$ $step$}
    \EndFor
    \State \Return {$GlobalBest$}
\EndFunction
\end{algorithmic}
\end{algorithm}

\begin{algorithm} 
\begin{algorithmic}[1] 
\caption{ABCO algorithm: exploitation step}\label{algo:algo_exploit}
\tiny
\Statex
    \For{$explt \gets 1$ to $N_{explt}$} \Comment{$Exploitation$ $step$}
    \For {$b $in$ p$}
    \If{$mode \gets "min"$}
    \State{$b.position \gets $$b.position$ $+$ $N_s$ $*$ $distance(b.position, min(b.position(1,k)))$ }\EndIf\Comment {move towards smallest solution out of k neighbours}
    \If{$mode \gets "max"$}
    \State{$b.position \gets $$b.position$ $+$ $N_s$ $*$ $distance(b.position, max(b.position(1,k)))$ }\EndIf\Comment {move towards largest solution out of k neighbours}
    \EndFor
    \EndFor\\
    
\end{algorithmic}
\end{algorithm}

\begin{algorithm} 
\begin{algorithmic}[1] 
\caption{ABCO algorithm: reproduction step}\label{algo:algo_repro}
\tiny
\Statex
    \Comment{$Reproduction$ $step$}
    \If{$mode \gets "min"$}
    \State{$List \gets  {SortedAscend(p)}$} \Comment {sort in ascending order}
    \EndIf 
    \If{$mode \gets "max"$}
    \State{$List \gets  {SortedDescend(p)}$} \Comment {sort in descending order}
    \EndIf
    
    \State{$p \gets $$the$ $Top$ $s$ $bacteria$ $in$ $p$}
    \For{$b \gets $$s+1$ $to$ $size$}
    \State{$p \gets  p.add(bmodified)$}  \Comment{Each modified b is the weighted sum average among the K neighbours of top s bacteria}
    \EndFor

    \If{$iter$ $\%$ $generationgap$ $==0$}
    
    \For{$b \gets $$s+1$ $to$ $size$}
     \If{$b.bestsolution$ $==$ $b.previousbestsolution$}
    \State{$unchangedBacteria$ ++} \Comment count number of unchanged bacteria
    \EndIf
     \If{(($unchangedBacteria \div size$ )*100) $>$ $unchangedThreshold$}
    \State{\Return {$GlobalBest$}} \Comment stop algorithm if the majority of bacteria solutions didn't change since previous check
    \Else 
    \State{$b.previousbestsolution$ $=$ $b.bestsolution$} \Comment update $b.previousbestsolution$
    \EndIf
    \EndFor
    \EndIf
    \State{$Update$ $GlobalBest$}
    
\end{algorithmic}
\end{algorithm}

The exploitation stage (Algorithm \ref{algo:algo_exploit}) is controlled by a tunable parameter $N_{explt}$, which dictates the number of times this stage is run. In the exploitation stage, the Euclidean distance between each bacterium and its neighbouring bacteria is calculated. A tuneable parameter $k$ is used to set the number of nearest neighbours that should be considered for each bacterium. The direction of each bacterium's movement towards the best solution amongst its $k$ neighbours is determined by whether the problem is a minimisation or maximisation problem.

The final stage of the ABCO algorithm is the reproduction stage. This stage is run once per iteration (Algorithm \ref{algo:algo_repro}). First, all the bacteria are sorted in the ascending or descending order depending on whether it is a minimisation or maximisation problem, respectively (Algorithm \ref{algo:algo_repro}, lines 1-6). The top $s$ bacteria are selected from the sorted list to proceed to the next generation, i.e. to be used in the next algorithm iteration (Algorithm \ref{algo:algo_repro}, line 7). The remaining bacteria to make up the population size ($size$--$s$) are generated using the weighted average of the positions of the $k$ neighbours of each of the top $size$--$s$ bacteria (Algorithm \ref{algo:algo_repro}, lines 8-10). After running for $iter$ iterations, the algorithm returns $GlobalBest$ -- the best solution found among all the bacteria in the population (Algorithm \ref{algo:algo_repro}, line 23). 

Depending on the application, an alternative stopping criterion can be activated through the $generationgap$ and $unchangedThreshold$ parameters to ensure timely completion of the algorithm (Algorithm \ref{algo:algo_repro}, lines 11-20). In particular, if the best bacteria solutions remain unchanged for $generationgap$ proportion of the total number of iterations $iter$, then the algorithm halts. For example, if $iter$=200 and $generationgap$=25\%, then every 50 (25\% of 200) iterations, the algorithm checks whether the current best solutions of bacteria remained unchanged compared to their best solutions found 50 iterations ago. The algorithm stops if the $unchangedThreshold$ proportion of the total population ($size$) did not change their solutions; otherwise, the algorithm continues to run until either the next checkpoint is triggered or the total number of iterations is up.

\section{Experiments}\label{Experimentssect}

To evaluate the proposed ABCO algorithm, its performance was compared to that of ACO \cite{dorigo1996} and PSO \cite{kennedy1995} over the widely used test functions detailed in Tables \ref{table:testfunctionformula} and \ref{table:testfunctions}. ACO and PSO were chosen as baselines in this study due to their popularity and code availability (the code for the previously proposed BCO algorithm \cite{niu2012} is not publicly available). The mealpy Python library was used to implement both the ten test functions and the two baseline algorithms. The error rate (the absolute difference between the found and true solutions; in this case, the best global minimum value found by the algorithms and the true global minimum value of the function) and runtime (the time it took the algorithms to output the result) were used as performance metrics, noting that there is typically a trade-off between the two metrics: shorter runtimes lead to poorer results. 

\begin{table}
\tiny
\caption{Test functions: equation}
\centering
\begin{tabular}{||p{1.1cm} p{7.1cm}||} 
 \hline
 Function Name & Equation \\ 
 \hline\hline
 Ackley  & $
f(x, y)=-20 \exp \left[-0.2 \sqrt{0.5\left(x^2+y^2\right)}\right]
-\exp [0.5(\cos 2 \pi x+\cos 2 \pi y)]+e+20
$\\
&\\
 Schaffer  &  $
f(x, y)=0.5+\frac{\sin ^2\left(x^2-y^2\right)-0.5} {\left[1+0.001\left(x^2+y^2\right)\right]^2}
$\\
&\\
 Rastrigin   & 
$f(x)=A n+\sum_{i=1}^n\left[x_i^2-A \cos \left(2 \pi x_i\right)\right]$
 \\
&\\

 Holder's Table   & 
$f(x, y)=-\left|\sin x \cos y \exp \left(\left|1-\frac{\sqrt{x^2+y^2}}{\pi}\right|\right)\right| $
\ \\ 
&\\
Rosenbrock & $f(x)=\sum_{i=1}^{n-1}\left[100(x_{i+1}-x_{i}^{2})^{2}+(1-x_{i})^{2}\right]$ \\
&\\
Sphere & $f(x)=\sum_{i=1}^{n}x_{i}^{2}$ \\
&\\
Booth & $f(x,y)=\left(x+2y-7\right)^{2}+\left(2x+y-5\right)^{2}$
 \\
&\\
Easom & $f(x,y)=-\cos(x)\cos(y)\exp\Bigl(-\Bigl((x-\pi)^{2}+(y-\pi)^{2}\Bigr)\Bigr)$
\\
&\\
Himmelblau & $f(x,y)= (x^{2}+y-11)^{2}+(x+y^{2}-7)^{2}$
 \\
&\\
&\\
Goldstein-price & $\begin{array}{c}{{f(x,y)=\left[1+\left(x+y+1\right)^{2}\left(19-14x+3x^{2}-14y+6x y+3y^{2}\right)\right]}}\\ {{\left[30+\left(2x-3y\right)^{2}\left(18-32x+12x^{2}+48y-36x y+27y^{2}\right)\right]}}\end{array}$
\\[1ex] 

\hline
\end{tabular}
\label{table:testfunctionformula}
\end{table}

\begin{table}
\tiny
\caption{Test functions: solution and search space}
\centering
\begin{tabular}{||c c c||} 
 \hline
 Function Name & Global minimum & Search space \\ [0.5ex] 
 \hline\hline
 Ackley & $f$(0,0)=0 & [-5, 5] \\
&&\\
 
 Schaffer & $f$(0, 0)=0 & [-100, 100]  \\
 &&\\
 Rastrigin & $f$=0 & [-5.12, 5.12] \\
 &&\\
 Holder's Table & $f$(8.05502, 9.66459)= -19.2085 & [-10, 10]\\ 
 &&\\
 Rosenbrock & $f$(1, 1)=0 & [-5, 10]\\
 &&\\
 Sphere & $f$(0) = 0 & [-100, 100]\\
 &&\\
 Booth & $f$(1, 3)=0 & [-10, 10] \\
 &&\\
 Easom & $f(\pi, \pi)$=-1 &[-100, 100]\\
 &&\\
 Himmelblau & $f$(3.0, 2.0)=0.0 & [-5, 5]\\
 &&\\
 Goldstein-price & $f$(0, -1)=3 & [-2, 2]

 \\ [1ex] 
 \hline
\end{tabular}
\label{table:testfunctions}
\end{table}

\begin{table}
\tiny
\caption{Parameter values of the ant colony optimisation (ACO) algorithm used in the experiments}
\centering
\begin{tabular}{||c c c||} 
 \hline
 Parameter & Description & Value \\ [0.5ex] 
 \hline\hline
  $sample_{count}$ & Number of Newly Generated Samples & 25 (5) \\
  $intent_{factor}$ &  Intensification Factor (Selection Pressure) & 0.5\\
  $zeta$ &  Deviation-Distance Ratio  & 1.0
 \\ [1ex] 
 \hline
\end{tabular}
\label{table:ACOParamDesc}
\end{table}

\begin{table}
\tiny
\caption{Parameter values of the particle swarm optimisation (PSO) algorithm used in the experiments}
\centering
\begin{tabular}{||c c c||} 
 \hline
 Parameter & Description & Value \\ [0.5ex] 
 \hline\hline
  $C_1$ & local coefficient  & 1.90 \\
  $C_2$ & global coefficient & 1.90\\
  $W_{min}$ &  Weight min of bird  & 0.4\\
 $W_{max}$ &Weight max of bird & 0.5
 \\ [1ex] 
 \hline
\end{tabular}
\label{table:PSOParamDesc}
\end{table}

\begin{table}
\tiny
\caption{Parameter values of the proposed adaptive bacteria colony optimisation (ABCO) algorithm used in the experiments}
\centering
\begin{tabular}{||p{1.1cm} p{6cm}||} 
 \hline
 Function & Configurations \\ [0.5ex] 
 \hline\hline
 Ackley & $N_s=1$ , $N_{explor}=4$, $N_{explt}=1$, $N_{tum}=1$, $e=0.05$, $s=0.8$, $k=2$\\
 &\\
 Holder's table & $N_s=1$, $N_{explor}=4$, $N_{explt}=1$, $N_{tum}=1$, $e=0.4$, $s=0.8$, $k=2$\\
 &\\
 Goldstein-price & $N_s=1$, $N_{explor}=4$, $N_{explt}=1$, $N_{tum}=1$, $e=0.05$, $s=0.8$, $k=2$\\
 &\\
 Easom & $N_s=1$, $N_{explor}=4$, $N_{explt}=1$, $N_{tum}=1$, $e=0.05$, $s=0.8$, $k=2$\\
 &\\
 Schaffer & $N_s=1$, $N_{explor}=4$, $N_{explt}=1$, $N_{tum}=1$, $e=0.3$, $s=0.8$, $k=2$\\
&\\
 Rastrigin & $N_s=1$, $N_{explor}=4$, $N_{explt}=1$, $N_{tum}=1$, $e=0.05$, $s=0.8$, $k=2$\\
 &\\
 Rosenbrock & $N_s=1$, $N_{explor}=4$, $N_{explt}=1$, $N_{tum}=1$, $e=0.4$, $s=0.8$, $k=2$\\
 &\\
 Booth & $N_s=1$, $N_{explor}=4$, $N_{explt}=1$, $N_{tum}=1$, $e=0.05$, $s=0.8$, $k=2$\\
 &\\
 Himmelblau & $N_s=1$, $N_{explor}=4$, $N_{explt}=1$, $N_{tum}=1$, $e=0.05$, $s=0.3$, $k=5$\\
 &\\
 Sphere & $N_s=1$, $N_{explor}=4$, $N_{explt}=1$, $N_{tum}=3$, $e=0.4$, $s=0.5$, $k=15$
 \\ [1ex] 
 \hline
\end{tabular}
\label{table:specsperfunction}
\end{table}

Three experiments were run to fully demonstrate how the three algorithms cope with this trade-off. The population size parameter of the algorithms was used to control the runtime and illustrate its impact on algorithms' accuracy. In the first and second experiments, the population size was set to a high number of 100 and a low number of 25, respectively, for all three algorithms. To stress-test the proposed ABCO algorithm in the third experiment, the population size for it was set to 25 (15 for the Sphere function to keep runtime comparable across the functions), while allowing the baseline ACO and PSO algorithms to run for longer with a population size of 100. All algorithms were run 50 times in each of the three experiments to demonstrate the degree of algorithms' reliability.

The parameter values set for ACO, PSO and the proposed ABCO algorithm are detailed in Tables \ref{table:ACOParamDesc}, \ref{table:PSOParamDesc} and \ref{table:specsperfunction}, respectively. Note that depending on the population size, 100 or 25, the $sample_{count}$ parameter of the ACO algorithm was set to 25 or 5, respectively, to allow correct running of the algorithm (Table \ref{table:ACOParamDesc}). While otherwise all ACO and PSO parameters were set to their default (overall best) values, some of ABCO parameters were tuned to the test functions to leverage the adaptive nature of the algorithm. In this study, given the simplicity of the test functions, the early stopping criterion was enabled by setting the $generationgap$ parameter to 25\% and the $unchangedThreshold$ parameter to 80\%. This means that the algorithm could stop after concluding only 25\%, 50\% or 75\% of the total number of iterations ($iter$) if the solutions of the 80\% of the total populations ($size$) remained unchanged compared to the solutions found at the end of the previous 25\% of the total number of iterations.

The experiments were run on a machine with the Windows Server 2016 operating system, two Intel(R) Xeon(R) Silver 4214R processors with a clock speed of 2.40 GHz and 2.39 GHz, respectively, and 64 GB of RAM.

\section{Results and Discussion}\label{FAT}
\begin{figure}
\centering
\includegraphics[width=14cm]{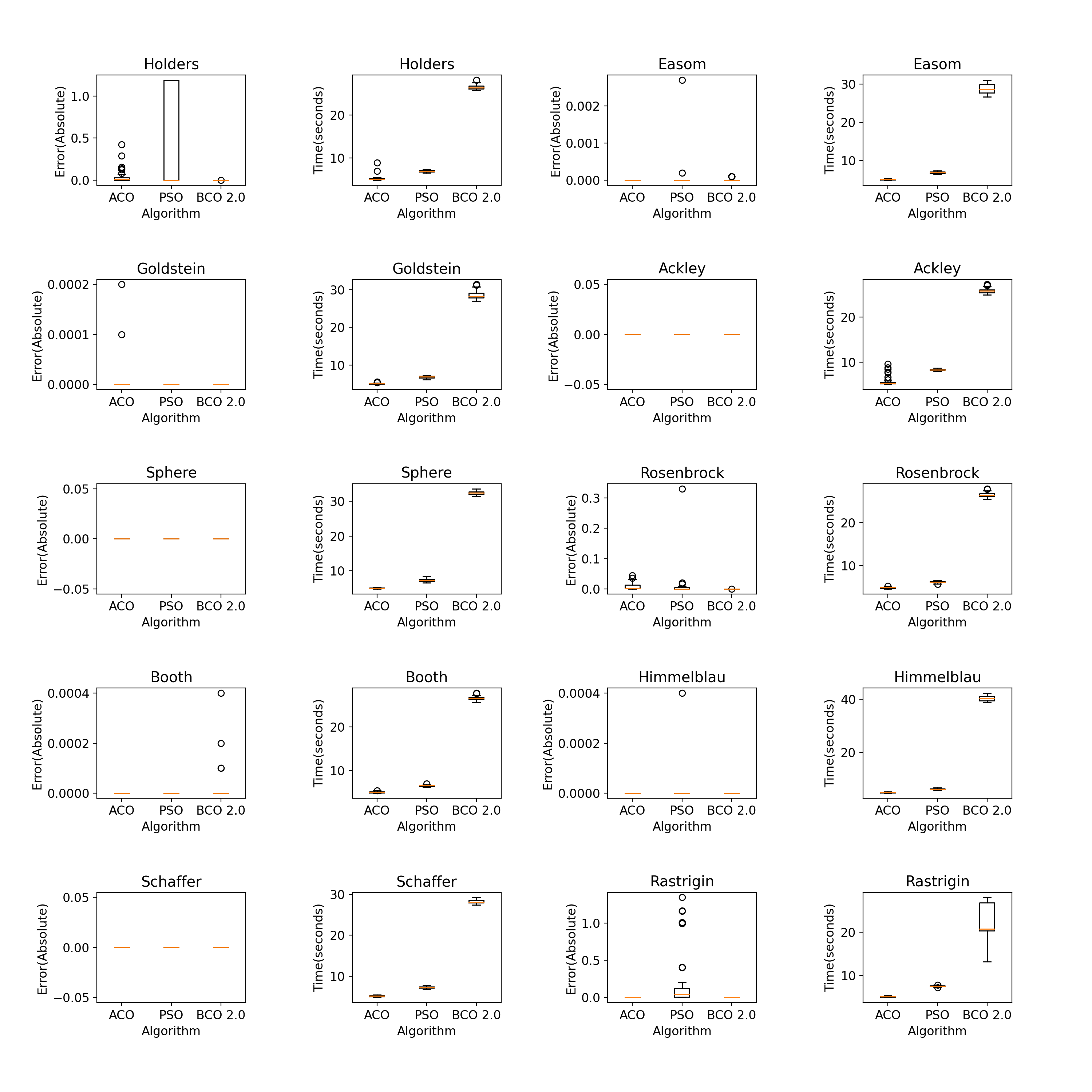}
\caption{Experiment 1 (population size of 100): error rate and runtime results of the ACO, PSO and ABCO algorithms over 50 runs for each of the ten test functions (error rates are shown in the first and third columns, while runtime -- in the second and fourth columns).}\label{fig:figure-popsize=100}
\end{figure}

The results of the first experiment (the population size set to 100 for all three algorithms) are illustrated in Fig. \ref{fig:figure-popsize=100}. The error rates (the absolute difference between the found and true solutions) achieved by the three algorithms in the first experiment for each of the ten test functions and  the runtimes (seconds) are detailed in Table \ref{table:datapop=100}. The table lists the best, worst and mean results, along with standard deviation, over 50 runs. It can be noticed from Fig. \ref{fig:figure-popsize=100} and Table \ref{table:datapop=100} that the proposed ABCO algorithm outputs more accurate and stable results than either ACO or PSO, or both, on 5 out of 10 test functions (Holder's table, Rosenbrock, Goldstein-price, Rastrigin and Himmelblau), while all three algorithms perform similarly accurate on all the other test functions. However, the better performance of ABCO in the first experiment is achieved at the cost of the runtime: ABCO takes longer to produce results for all ten test functions compared to ACO and PSO when set to run with a high population size.

\begin{figure}
\centering
\includegraphics[width=14cm]{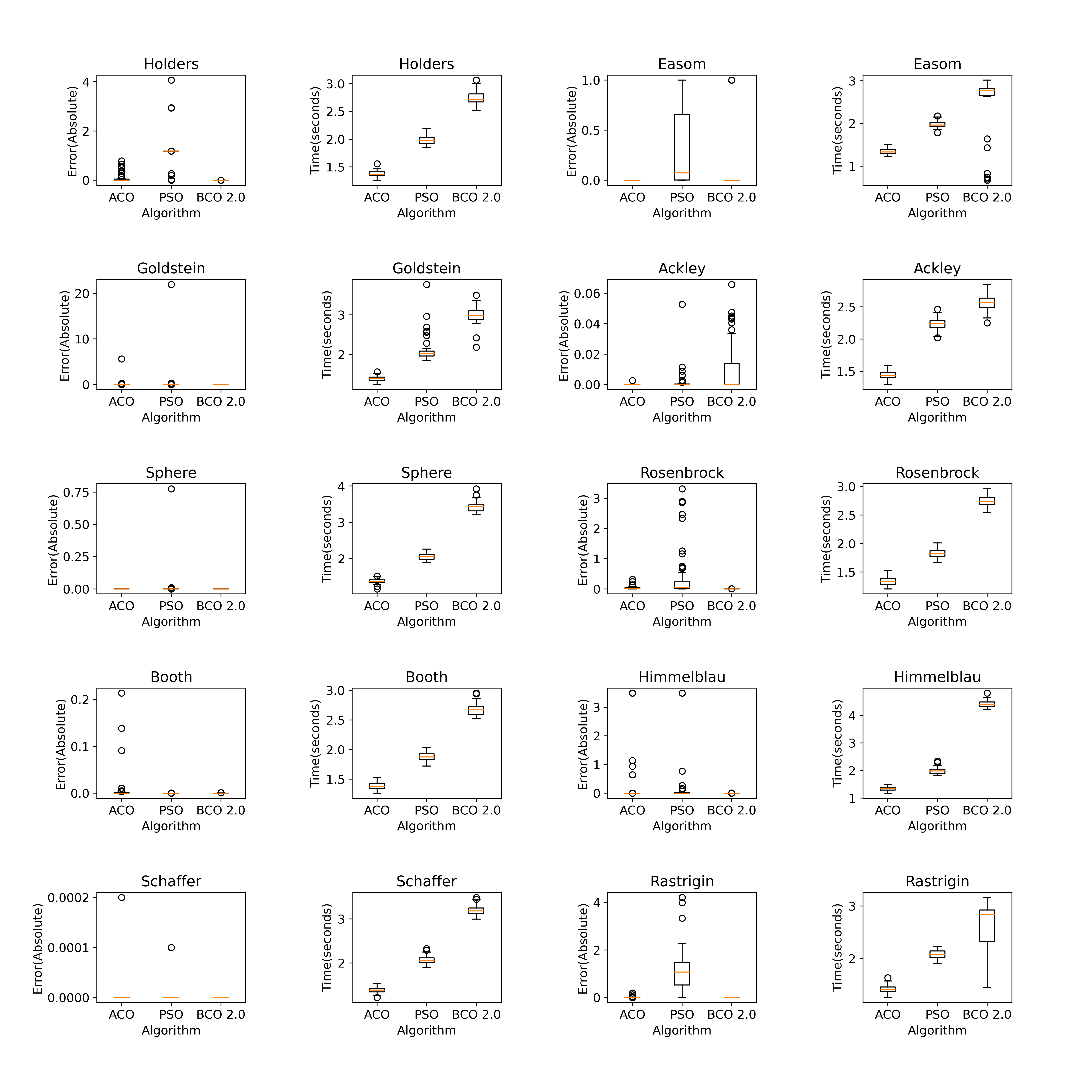}\\
\caption{Experiment 2 (population size of 25): error rate and runtime results of the ACO, PSO and ABCO algorithms over 50 runs for each of the ten test functions.} \label{fig:figure-popsize=25}
\end{figure}

\begin{figure}
\centering
\includegraphics[width=14cm]{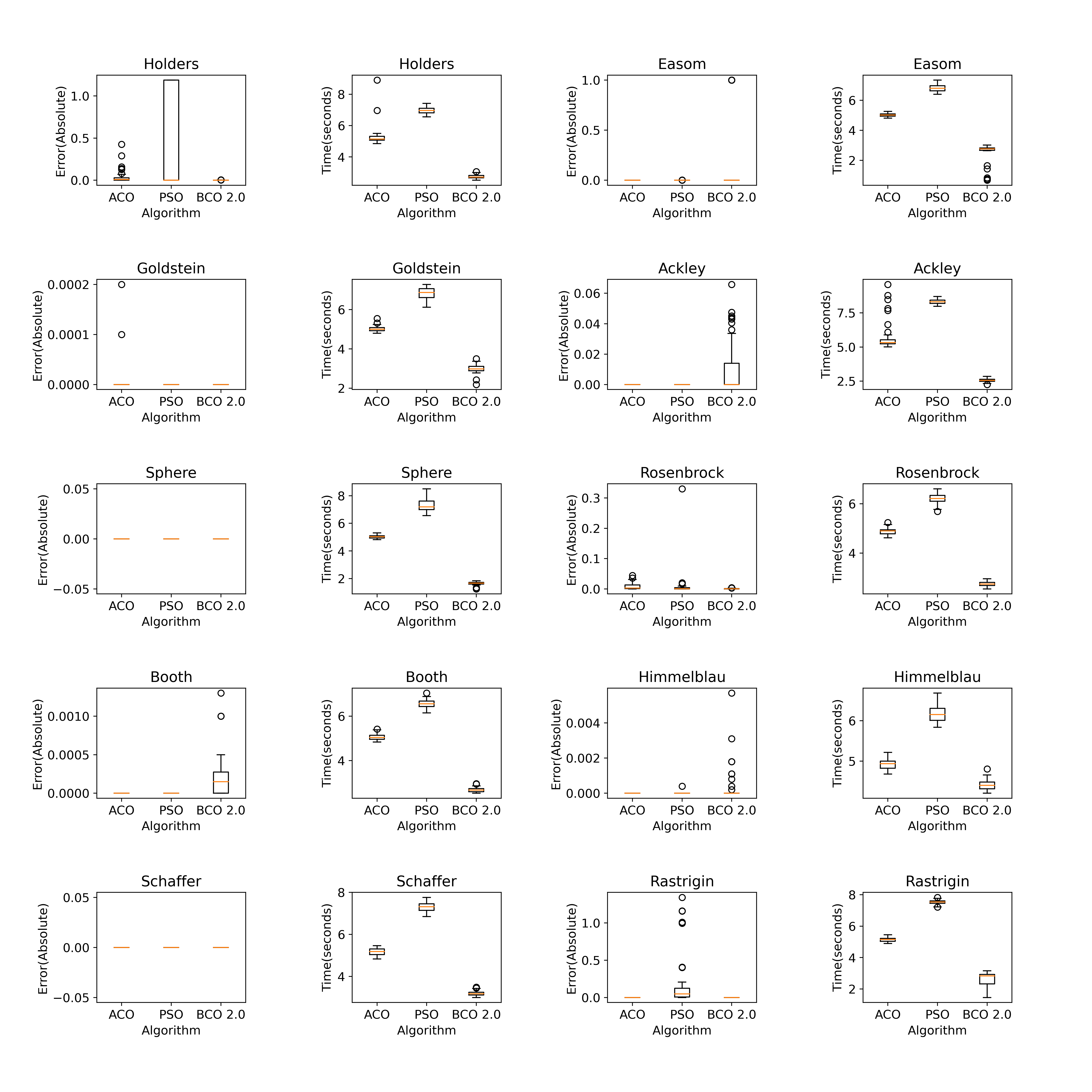}\\
\caption{Experiment 3: error rate and runtime results of the ACO, PSO and ABCO algorithms over 50 runs for each of the ten test functions. ABCO population size set to 25 (15 for the Sphere function); ACO and PSO population sizes set to 100.} \label{figure-ACOPSO=100-BCOpopsize=25-sphere-15}
\end{figure}


While still running slower (by at most 2 seconds) than ACO and PSO in the second experiment, where the population size was set to 25 for all three algorithms, ABCO again achieves more accurate and stable results than the other two algorithms, now on 7 out of 10 test functions: Holder's table, Goldstein-price, Booth, Rosenbrock, Himmelblaue, Schaffer and Rastrigin (see Fig. \ref{fig:figure-popsize=25} and Table \ref{table:dataExperiment2}). This experiment demonstrates that unlike ACO and PSO, ABCO can achieve accurate and stable results with a small population owing to the algorithm's exploration stage, which enables a limited population to effectively scan the search space.

Given the strong performance of ABCO in the second experiment, the third experiment was conducted to compare the performance of the "light" version of the proposed ABCO algorithm set with a population size of 25 (15 for the Sphere function to keep runtimes comparable across the board) with the "heavy" but more accurate versions of ACO and PSO, both set with a population size of 100. It can be noticed from Fig. \ref{figure-ACOPSO=100-BCOpopsize=25-sphere-15} and Table \ref{table:data-Experiment3-spherepop=15}, which detail the results of the three algorithms in the third experiment, that the proposed ABCO algorithm outputs results much faster compared to ACO and PSO (up to 6 seconds across all 10 test functions), while maintaining competitive performance, with ABCO outperforming PSO and ACO on the Rastrigin, Holder's table, Rosenbrock and Goldstein-price test functions.


\begin{table}
\tiny
\caption{Experiment 1 (population size of 100): error rate and runtimes (seconds) of the ACO, PSO and ABCO algorithms over 10 test functions. The best results are highlighted in bold.}
\begin{tabular}{|| p{1.1cm}p{0.5cm}|p{0.7cm}p{0.7cm}p{0.9cm} |p{0.7cm}p{0.7cm}p{0.9cm}||}
\hline
     \multirow{2}{*}{Test Functions} &
     \multicolumn{1}{l}{} &
      \multicolumn{3}{|c|}{Error rate} &
      \multicolumn{3}{|c||}{Runtime} \\ [7pt]
       &{}&{ACO}& {PSO} & {ABCO} &{ACO}& {PSO} & {ABCO} \\[7pt]
\hline

Rastrigrin & Best &  0.0  & 0.0  & 0.0
& \textbf{4.90122}   & 7.21806  &13.20239
\\
& Worst & \textbf{0.0} & 1.3428 & \textbf{0.0 }
& \textbf{5.44912} & 7.83003 & 28.03324 
\\
& Mean &\textbf{ 0.0 }&  0.18901 & \textbf{0.0}
& \textbf{5.1384 }&  7.52436 & 23.22504
\\
& std  & \textbf{0.0} & 0.35363 & \textbf{0.0 }
& \textbf{0.12558 }& 0.1403 & 3.59996 
\\
&&&&&&& \\
\hline
Ackley & Best &  0.0  & 0.0  & 0.0
& \textbf{5.01675}  & 7.97859  &24.90557
\\
& Worst & 0.0 & 0.0 & 0.0 
& 9.58464 & \textbf{8.69868} & 27.24048 
\\
& Mean & 0.0 &  0.0 & 0.0
&\textbf{5.68586}&  8.32703 & 25.82983
\\
& std  & 0.0 & 0.0 & 0.0 
& 0.99226 &\textbf{0.17261}& 0.52274 
\\
&&&&&&& \\
\hline
Schaffer & Best &  0.0  & 0.0  & 0.0
& \textbf{4.83171}  & 6.84942  &27.39834
\\
& Worst & 0.0 & 0.0 & 0.0 
& \textbf{5.46129} & 7.76242 & 29.22421 
\\
& Mean & 0.0 &  0.0 & 0.0
& \textbf{5.17267} &  7.31111 & 28.20971
\\
& std  & 0.0 & 0.0 & 0.0 
& \textbf{0.15893} & 0.20308 & 0.45377 
\\
&&&&&&& \\
\hline
Holder's table & Best &  -0.0  & -0.0  & 0.0
& \textbf{4.84834}  & 6.55658  &25.70612
\\
& Worst & 0.4255 & 1.1878 & \textbf{0.0013}
& 8.89895 & \textbf{7.41715} & 28.06256 
\\
& Mean & 0.03619 &  0.3801 & \textbf{0.00034}
& \textbf{5.2776} &  6.9656 & 26.43957
\\
& std  & 0.07611 & 0.55408 & \textbf{0.00032}
& 0.59543 & \textbf{0.20338} & 0.52784 
\\
&&&&&&& \\
\hline
Rosenbrock & Best &  0.0  & 0.0  & 0.0
& \textbf{4.6171 }  & 5.68405  &25.36106
\\
& Worst & 0.0441 & 0.3304 & \textbf{0.0001}
& \textbf{5.23933} & 6.6063 & 27.8292 
\\
& Mean & 0.00837 &  0.0091 & \textbf{0.0}
& \textbf{4.88149} &  6.20991 & 26.39082
\\
& std  & \textbf{0.01044} & 0.04609 & 1e-05 
& \textbf{0.12486} & 0.18718 & 0.48252 
\\
&&&&&&& \\
\hline
Sphere & Best &  0.0  & 0.0  & 0.0
& \textbf{4.8169}   & 6.56431  &31.48605
\\
& Worst & 0.0 & 0.0 & 0.0 
& \textbf{5.30754} & 8.50461 & 33.62666 
\\
& Mean & 0.0 &  0.0 & 0.0
& \textbf{5.02894} &  7.31703 & 32.41007
\\
& std  & 0.0 & 0.0 & 0.0 
& \textbf{0.10995} & 0.46319 & 0.53289 
\\
&&&&&&& \\
\hline
Booth & Best &  0.0  & 0.0  & 0.0
& \textbf{4.83417}   & 6.15052  &25.64179
\\
& Worst & \textbf{0.0} & \textbf{0.0} & 0.0004 
& \textbf{5.41818} & 7.04274 & 27.71236 
\\
& Mean & \textbf{0.0} &  \textbf{0.0} & 2e-05
& \textbf{5.05434} &  6.54652 & 26.537
\\
& std  & \textbf{0.0} & \textbf{0.0} & 6e-05 
& \textbf{0.13178} & 0.19 & 0.42208 
\\
&&&&&&& \\
\hline
Easom & Best &  0.0  & 0.0  & 0.0
& \textbf{4.80775}   & 6.3956  &26.70998
\\
& Worst & \textbf{0.0} & 0.0027 & 0.0001 
& \textbf{5.25131} & 7.34018 & 31.07061 
\\
& Mean & \textbf{0.0}&  6e-05 & 2e-05
& \textbf{5.01769} &  6.80449 & 28.85576
\\
& std  & \textbf{0.0} & 0.00038 & 4e-05 
& \textbf{0.09671} & 0.22745 & 1.26279 
\\
&&&&&&& \\
\hline
Himmelblau & Best &  0.0  & 0.0  & 0.0
& \textbf{4.68033}   & 5.83753  &38.75495
\\
& Worst & \textbf{0.0} & 0.0004 & \textbf{0.0}
& \textbf{5.21528} & 6.68703 & 42.24242 
\\
& Mean & \textbf{0.0} &  1e-05 & \textbf{0.0}
& \textbf{4.9218} &  6.17414 & 40.22662
\\
& std  & \textbf{0.0} & 6e-05 & \textbf{0.0}
& \textbf{0.11642} & 0.20182 & 0.9759 
\\
&&&&&&& \\
\hline
Goldstein-price & Best &  0.0  & -0.0  & 0.0
& \textbf{4.78965}   & 6.11902  &27.03221
\\
& Worst & 0.0002 & \textbf{0.0} & \textbf{0.0}
& \textbf{5.53901} & 7.27436 & 31.40191 
\\
& Mean & 1e-05 &   \textbf{0.0} & \textbf{0.0}
& \textbf{5.01172} &  6.81497 & 28.59377
\\
& std  & 3e-05 & \textbf{0.0} & \textbf{0.0}
& \textbf{0.13575} & 0.28134 & 1.06717 
\\
&&&&&&& 
\\[1ex]
\hline
\end{tabular}
\label{table:datapop=100}
\end{table}

\begin{table}
\tiny
\caption{Experiment 2 (population size of 25): error rate and runtimes (seconds) of the ACO, PSO and ABCO algorithms over 10 test functions. The best results are highlighted in bold.}
\begin{tabular}{|| p{1.1cm}p{0.5cm}|p{0.7cm}p{0.7cm}p{0.9cm} |p{0.7cm}p{0.7cm}p{0.9cm}||}
\hline
     \multirow{2}{*}{Test Functions} &
     \multicolumn{1}{l}{} &
      \multicolumn{3}{|c|}{Error rate} &
      \multicolumn{3}{|c||}{Runtime} \\ [7pt] 
       &{}&{ACO}& {PSO} & {ABCO} &{ACO}& {PSO} & {ABCO} \\[7pt]
\hline

Rastrigrin & Best &  \textbf{0.0 } & 0.0069  & \textbf{0.0 }
& \textbf{1.26234}  & 1.91031  &1.45581
\\
& Worst & 0.192 & 4.2138 & \textbf{0.0}
& \textbf{1.63744} & 2.23284 & 3.16193 
\\
& Mean & 0.00775 &  1.20253 & \textbf{0.0}
& \textbf{1.42269} &  2.08716 & 2.67265
\\
& std  & 0.0314 & 0.90964 & \textbf{0.0}
& \textbf{0.07086} & 0.08218 & 0.38895 
\\
&&&&&&& \\
\hline
Ackley & Best &  0.0  & 0.0  & 0.0
& \textbf{1.29401}   & 2.0187  &2.25159
\\
& Worst & \textbf{0.0026} & 0.0528 & 0.0658 
& \textbf{1.59049} & 2.46099 & 2.84647 
\\
& Mean & 5e-05 &  0.00189 & 0.01048
& \textbf{1.43987} &  2.23109 & 2.56472
\\
& std  & \textbf{0.00036} & 0.00759 & 0.01786 
& \textbf{0.065} & 0.09403 & 0.11804 
\\
&&&&&&& \\
\hline
Schaffer & Best &  0.0  & 0.0  & 0.0
& \textbf{1.21655}   & 1.89161  &2.99482
\\
& Worst & 0.0002 & 0.0001 & \textbf{0.0}
& \textbf{1.5381} & 2.32517 & 3.48616 
\\
& Mean & 0.0 &  0.0 & 0.0
& \textbf{1.38609} &  2.07234 & 3.18917
\\
& std  & 3e-05 & 2e-05 & \textbf{0.0}
& \textbf{0.06332} & 0.08889 & 0.11406 
\\
&&&&&&& \\
\hline
Holder's table & Best &  -0.0  & -0.0  & 0.0
& \textbf{1.25995}  & 1.84901  &2.51224
\\
& Worst & 0.7848 & 4.0683 & \textbf{0.0057}
& \textbf{1.55559} & 2.19266 & 3.0637 
\\
& Mean & 0.09172 &  1.2303 & \textbf{0.0009}
& \textbf{1.37879} &  1.98343 & 2.73873
\\
& std  & 0.18256 & 0.76432 & \textbf{0.00103}
&\textbf{0.05535} & 0.07868 & 0.10668 
\\
&&&&&&& \\
\hline
Rosenbrock & Best &  0.0  & 0.0  & 0.0
& \textbf{1.20592}   & 1.66736  &2.54578
\\
& Worst & 0.316 & 3.3156 & \textbf{0.004}
& \textbf{1.53184} & 2.01191 & 2.95972 
\\
& Mean & 0.03266 &  0.42083 & \textbf{0.00062}
& \textbf{1.34905} &  1.83004 & 2.74004
\\
& std  & 0.05786 & 0.84013 & \textbf{0.00091}
& 0.07073 & \textbf{0.06695} & 0.08475 
\\
&&&&&&& \\
\hline
Sphere & Best &  0.0  & 0.0  & 0.0
& \textbf{1.16927}   & 1.9043  &3.20315
\\
& Worst & \textbf{0.0} & 0.7761 & \textbf{0.0}
& \textbf{1.52387} & 2.26623 & 3.92375 
\\
& Mean & \textbf{0.0} &  0.01581 & \textbf{0.0}
& \textbf{1.378} &  2.0558 & 3.42016
\\
& std  &\textbf{0.0} & 0.10862 & \textbf{0.0}
& \textbf{0.06299} & 0.0853 & 0.13995 
\\
&&&&&&& \\
\hline
Booth & Best &  0.0  & 0.0  & 0.0
& \textbf{1.26405}   & 1.72305  &2.52645
\\
& Worst & 0.2136 & \textbf{0.0005} & 0.0013 
& \textbf{1.53101} & 2.03541 & 2.95394 
\\
& Mean & 0.00982 &  \textbf{2e-05} & 0.00021
& \textbf{1.38475} &  1.87731 & 2.68166
\\
& std  & 0.03703 & \textbf{8e-05} & 0.00027 
& \textbf{0.06071} & 0.07113 & 0.10208 
\\
&&&&&&& \\
\hline
Easom & Best &  0.0  & 0.0  & 0.0
& \textbf{1.22374}   & 1.78429  &0.67008
\\
& Worst & \textbf{0.0} & 1.0 & 1.0 
& \textbf{1.51219} & 2.17442 & 3.01521 
\\
& Mean & \textbf{0.0} &  0.3124 & 0.22011
& \textbf{1.34692} &  1.97558 & 2.3723
\\
& std  & \textbf{0.0} & 0.39776 & 0.41419 
& \textbf{0.06162} & 0.0757 & 0.80789 
\\
&&&&&&& \\
\hline
Himmelblau & Best &  0.0  & 0.0  & 0.0
& \textbf{1.17903}   & 1.83009  &4.2047
\\
& Worst & 3.4933 & 3.4937 & \textbf{0.0057}
& \textbf{1.48669} & 2.34058 & 4.80696 
\\
& Mean & 0.264 &  0.51781 & \textbf{0.00032}
& \textbf{1.34995} &  1.99158 & 4.41881
\\
& std  & 0.84517 & 1.2061 & \textbf{0.00097}
& \textbf{0.06865} & 0.10931 & 0.12677 
\\
&&&&&&& \\
\hline
Goldstein-price & Best &  0.0  & 0.0  & 0.0
& \textbf{1.23966 }  & 1.84732  &2.18337
\\
& Worst & 5.6093 & 21.9654 & \textbf{0.0}
& \textbf{1.56182} & 3.76966 & 3.49788 
\\
& Mean & 0.11912 &  0.44577 & \textbf{0.0}
& \textbf{1.38769} &  2.11289 & 2.99524
\\
& std  & 0.78536 & 3.07454 & \textbf{0.0}
& \textbf{0.06641} & 0.31889 & 0.21777 
\\
&&&&&&&
\\[1ex]
\hline
\end{tabular}
\label{table:dataExperiment2}
\end{table}

Several observations can be made based on the results obtained in the three experiments. First, the proposed ABCO algorithm can produce consistent and accurate results when the runtime is not crucial. The first and second experiments support this observation as ABCO outperforms PSO and ACO on 5 and 7 out of 10 test functions, respectively. Another observation that can be made from the results of the third experiment is the capability of the proposed ABCO algorithm to provide competitive results in a much shorter time and utilising fewer computational resources compared to ACO and PSO. In particular, ABCO completed up to 6 seconds faster than ACO and PSO on all test functions while still producing more accurate and stable results on 4 out of 10 test functions and achieving similar accuracy on the remaining test functions. Overall, it can be concluded that the proposed ABCO algorithm can be configured to outperform the ACO and PSO algorithms depending on the application requirements.

\begin{table}
\tiny
\caption{Experiment 3: error rate and runtimes (seconds) of the ACO, PSO and ABCO algorithms over 10 test functions. ABCO population size set to 25 (15 for the Sphere function); ACO and PSO population sizes set to 100. The best results are highlighted in bold.}
\begin{tabular}{|| p{1.1cm}p{0.5cm}|p{0.7cm}p{0.7cm}p{0.9cm} |p{0.7cm}p{0.7cm}p{0.9cm}||}
\hline
     \multirow{2}{*}{Test Functions} &
     \multicolumn{1}{l}{} &
      \multicolumn{3}{|c|}{Error rate} &
      \multicolumn{3}{|c||}{Runtime} \\ [7pt] 
       &{}&{ACO}& {PSO} & {ABCO} &{ACO}& {PSO} & {ABCO} \\[7pt]
\hline
Rastrigrin & Best &  0.0  & 0.0  & 0.0
& 4.90122   & 7.21806  & \textbf{1.45581}
\\
& Worst & \textbf{0.0} & 1.3428 & \textbf{0.0}
& 5.44912 & 7.83003 & \textbf{3.16193}
\\
& Mean & \textbf{0.0} &  0.18901 & \textbf{0.0}
& 5.1384 &  7.52436 & \textbf{2.67265}
\\
& std  & \textbf{0.0} & 0.35363 & \textbf{0.0} 
& \textbf{0.12558} & 0.1403 & 0.38895 
\\
&&&&&&& \\
\hline
Ackley & Best &  0.0  & 0.0  & 0.0
& 5.01675   & 7.97859  &\textbf{2.25159}
\\
& Worst & \textbf{0.0} & \textbf{0.0} & 0.0658 
& 9.58464 & 8.69868 & \textbf{2.84647}
\\
& Mean & \textbf{0.0} &  \textbf{0.0} & 0.01048
& 5.68586 &  8.32703 &\textbf{2.56472}
\\
& std  & \textbf{0.0} & \textbf{0.0} & 0.01786 
& 0.99226 & 0.17261 & \textbf{0.11804}
\\
&&&&&&& \\
\hline
Schaffer & Best &  0.0  & 0.0  & 0.0
& 4.83171   & 6.84942  & \textbf{2.99482}
\\
& Worst & 0.0 & 0.0 & 0.0 
& 5.46129 & 7.76242 & \textbf{3.48616}
\\
& Mean & 0.0 &  0.0 & 0.0
& 5.17267 &  7.31111 & \textbf{3.18917}
\\
& std  & 0.0 & 0.0 & 0.0 
& 0.15893 & 0.20308 & \textbf{0.11406}
\\
&&&&&&& \\
\hline
Holder's table & Best &  -0.0  & -0.0  & 0.0
& 4.84834   & 6.55658  &\textbf{2.51224}
\\
& Worst & 0.4255 & 1.1878 & \textbf{0.0057}
& 8.89895 & 7.41715 & \textbf{3.0637}
\\
& Mean & 0.03619 &  0.3801 & \textbf{0.0009}
& 5.2776 &  6.9656 & \textbf{2.73873}
\\
& std  & 0.07611 & 0.55408 & \textbf{0.00103}
& 0.59543 & 0.20338 & \textbf{0.10668}
\\
&&&&&&& \\
\hline
Rosenbrock & Best &  0.0  & 0.0  & 0.0
& 4.6171   & 5.68405  & \textbf{2.54578}
\\
& Worst & 0.0441 & 0.3304 & \textbf{0.004} 
& 5.23933 & 6.6063 & \textbf{2.95972} 
\\
& Mean & 0.00837 &  0.0091 & \textbf{0.00062}
& 4.88149 &  6.20991 & \textbf{2.74004}
\\
& std  & 0.01044 & 0.04609 & \textbf{0.00091} 
& 0.12486 & 0.18718 & \textbf{0.08475}
\\
&&&&&&& \\
\hline
Sphere & Best &  0.0  & 0.0  & 0.0
& 4.8169   & 6.56431  & \textbf{1.25133}
\\
& Worst & 0.0 & 0.0 & 0.0 
& 5.30754 & 8.50461 & \textbf{1.84551}
\\
& Mean & 0.0 &  0.0 & 0.0
& 5.02894 &  7.31703 & \textbf{1.63193}
\\
& std  & 0.0 & 0.0 & 0.0 
& \textbf{0.10995} & 0.46319 & 0.14183 
\\
&&&&&&& \\
\hline
Booth & Best &  0.0  & 0.0  & 0.0
& 4.83417   & 6.15052  &\textbf{2.52645}
\\
& Worst & \textbf{0.0} & \textbf{0.0} & 0.0013 
& 5.41818 & 7.04274 & \textbf{2.95394}
\\
& Mean & \textbf{0.0} &  \textbf{0.0} & 0.00021
& 5.05434 &  6.54652 & \textbf{2.68166}
\\
& std  & \textbf{0.0} & \textbf{0.0} & 0.00027 
& 0.13178 & 0.19 & \textbf{0.10208}
\\
&&&&&&& \\
\hline
Easom & Best &  0.0  & 0.0  & 0.0
& 4.80775   & 6.3956  & \textbf{0.67008}
\\
& Worst & \textbf{0.0} & 0.0027 & 1.0 
& 5.25131 & 7.34018 & \textbf{3.01521}
\\
& Mean & \textbf{0.0} &  6e-05 & 0.22011
& 5.01769 &  6.80449 & \textbf{2.3723}
\\
& std  & \textbf{0.0} & 0.00038 & 0.41419 
& \textbf{0.09671} & 0.22745 & 0.80789 
\\
&&&&&&& \\
\hline
Himmelblau & Best &  0.0  & 0.0  & 0.0
& 4.68033   & 5.83753  & \textbf{4.2047}
\\
& Worst & \textbf{0.0} & 0.0004 & 0.0057 
& 5.21528 & 6.68703 & \textbf{4.80696}
\\
& Mean & \textbf{0.0} &  1e-05 & 0.00032
& 4.9218 &  6.17414 & \textbf{4.41881}
\\
& std  & \textbf{0.0} & 6e-05 & 0.00097 
& \textbf{0.11642} & 0.20182 & 0.12677 
\\
&&&&&&& \\
\hline
Goldstein-price & Best &  0.0  & -0.0  & 0.0
& 4.78965   & 6.11902  & \textbf{2.18337}
\\
& Worst & 0.0002 & \textbf{0.0} & \textbf{0.0} 
& 5.53901 & 7.27436 & \textbf{3.49788}
\\
& Mean & 1e-05 &  \textbf{0.0} & \textbf{0.0}
& 5.01172 &  6.81497 & \textbf{2.99524}
\\
& std  & 3e-05 & \textbf{0.0} & \textbf{0.0} 
& \textbf{0.13575} & 0.28134 & 0.21777 
\\
&&&&&&&
\\[1ex]
\hline
\end{tabular}
\label{table:data-Experiment3-spherepop=15}
\end{table}

\clearpage

\section{Conclusion}\label{conclusion}
This paper introduced a new optimisation algorithm called Adaptive Bacterial Colony Optimisation (ABCO) algorithm. The unique feature of ABCO compared to existing optimisation algorithms is its ability to balance exploration and exploitation behaviours when searching the solution space. The experimental results demonstrated that the proposed algorithm finds optimal solutions much faster than other optimisation algorithms such as ACO and PSO. In the future, we plan to evaluate the performance of the proposed algorithm on real-life problems and further refine the algorithm by experimenting with different reproduction strategies, in addition to the current solution of producing new offsprings by taking weighted average of the best performing individuals.

\bibliography{sn-bibliography}

\end{document}